\documentclass[10pt, a4paper]{article}

\usepackage{lrec2026} 

\usepackage{latexsym}
\usepackage[T1]{fontenc}

\usepackage[utf8]{inputenc}

\usepackage{microtype}
\usepackage{caption}
\usepackage{subcaption}

\usepackage{siunitx}

\usepackage{nicefrac}
\usepackage{algorithm}
\usepackage[noend]{algpseudocode}
\usepackage{float}

\DeclareMathAlphabet{\mathbcal}{OMS}{cmsy}{b}{n}
\usepackage{amsmath}
\usepackage{mathrsfs}
\usepackage{comment}

\usepackage{bm}
\usepackage{bbm}
\usepackage{amssymb}
\usepackage{paralist}
\usepackage{xcolor}
\usepackage{graphicx}

\definecolor{formalshade}{rgb}{0.95,0.95,1}

\newenvironment{formal}{%
  \MakeFramed{\advance\hsize-\width\FrameRestore}%
  \noindent\hspace{-4.55pt}
  \begin{adjustwidth}{4pt}{7pt}%
  
}
{
  \end{adjustwidth}\endMakeFramed%
}

\usepackage{colortbl}
\usepackage{booktabs}
\usepackage{framed}
\usepackage[strict]{changepage}

\usepackage{afterpage}

\usepackage{array}
\usepackage{tablefootnote}
\newcolumntype{S}{>{\raggedright}r@{/}l}

\title{Blind to the Human Touch:\\Overlap Bias in LLM-Based Summary Evaluation}

\name{Jiangnan Fang{$^1$}, Cheng-Tse Liu{$^2$}, Hanieh Deilamsalehy{$^3$}, Nesreen K. Ahmed{$^4$},\\{\bf \large Puneet Mathur{$^5$}, Nedim Lipka{$^6$}, Franck Dernoncourt{$^7$}, Ryan A. Rossi{$^8$}}}

\address{{$^{1,2}$}Independent Researchers, {$^{3,5,6,7,8}$}Adobe Research, {$^4$}Cisco Research \\
         \{{$^1$}jfang53, {$^2$}cliu282\}@ucsc.edu }

\abstract{
Large language model (LLM) judges have often been used alongside traditional, algorithm-based metrics for tasks like summarization because they better capture semantic information, are better at reasoning, and are more robust to paraphrasing. However, LLM judges show biases for length and order among others, and are vulnerable to various adversarial input prompts. While recent studies have looked into these biases, few have analyzed them at a more granular level in relation to a well-defined overlap metric. In this work we provide an LLM judge bias analysis as a function of overlap with human-written responses in the domain of summarization. We test 9 recent LLMs with parameter counts ranging from 1 billion to 12 billion, including variants of Gemma 3 and LLaMA 3. We find that LLM judges increasingly prefer summaries generated by other LLMs over those written by humans as the similarities (as measured by ROUGE and BLEU) between the judged summaries decrease, and this pattern extends to all but one model tested, and exists regardless of the models' own position biases. Additionally, we find that models struggle to judge even summaries with limited overlaps, suggesting that LLM-as-a-judge in the summary domain should rely on techniques beyond a simple comparison.
 \\ \newline \Keywords{LLMs, summarization} }

\begin{document}

\maketitleabstract

As large language models (LLMs) continue to improve in their capabilities, LLM-as-a-judge has emerged as a method of automating evaluation. Compared to traditional overlap-based metrics, LLM judges better capture semantic content of texts and are more robust to paraphrasing. As models can leverage reasoning capabilities gained from training, they also enable evaluation that don't rely on reference texts that can be expensive to obtain \cite{wmt24}. However, bias remains an issue in the LLM-as-a-judge paradigm. Biases of judge LLMs reveal particular tendencies learned through LLM's training, and these tendencies displayed for one task (position bias, for example) extend to other areas and input domains \cite{tian2025identifyingmitigatingpositionbias} especially for zero-shot tasks as they rely on the internal knowledge of a model. Understanding these bias patterns is crucial in evaluating LLM performance and informing future training or design practices. While previous work has assessed how well LM decisions correlate with human judgments \cite{goyal2023newssummarizationevaluationera, zhu2025judgelmfinetunedlargelanguage}, including at different levels of quality as rated by humans \cite{shen-etal-2023-large}, the present work aims to look at correlations in a much more granular level in terms of levels of overlap as measured by n-gram metrics.

In this work, we study the following research questions:
(1) How does similarity as measured by n-gram overlap metrics (e.g. ROUGE, BLEU) correlate with LLM judgments, and is such judgment vs. similarity pattern observed for models of different sizes and types? (2) How does presentation order (position bias) interact with the judgment vs. similarity pattern?

After extensive experimentation, we make the following contributions and findings:

\begin{itemize}
\item
We collect a benchmark dataset containing 6,744 LLM summaries of a filtered subset of WikiSum and CNN\_DailyMail datasets and over 94,000 LLM judgments between human and machine-generated summaries.

\item
LLM judges' preference toward their own answers is (1) more prominent when generated summaries have fewer n-gram overlaps with the human-written summaries, \textbf{(2) exists even towards summaries by small models, and (3) a large difference is needed for such preference to show.} (See Figure~\ref{fig:stacked-gt-vs-gen} where the proportion of ``generated'' choices grows towards the left of individual results).

\item
Position bias is more prominent when the generated summaries are more like human-written summaries. In addition, models with more parameters tend to prefer the last-presented summaries while models with fewer parameters prefer the first-presented summaries. We note that the bias for generated summaries as described above persists for variously sized models we have tested, regardless of types of position bias. (See Figure~\ref{fig:stacked-sum-vs-eval} for further breakdowns for each evaluator and generator model.)

\item
We conduct a large-scale systematic study consisting of 9 LLMs, with parameter counts ranging from 1 billion to 12 billion, of judge LLM bias as a function vs. the degree of overlap between machine and human summaries.

\end{itemize}

\section{Related Works}\label{sec:related}

Recent studies in LLM judges include various evaluation datasets, frameworks, and prompting methods. \citet{zhu2025judgelmfinetunedlargelanguage} built a dataset for fine-tuning LLM judges and proposes a scalable framework for open-ended LLM judging tasks. \citet{longlamp} developed a benchmark dataset for evaluating personalized long text generation. \citet{hashemi-etal-2024-llm} develops a weighing framework for combining LLM responses to items in a rubric, and G-Eval in \citet{liu-etal-2023-g} uses a combination of chain-of-thought prompting and form-filling to improve a judge LLM's alignment with human preferences in summarization and text generation.

LLM judges exhibit content-level biases, i.e. they are biased towards textual content not related to the assigned evaluation task, or they ignore textual content that are related to evaluation. \citet{chen-etal-2024-humans} shows that LLM judges may overlook factual errors, and show gender, authority, and beauty biases. \citet{fu-etal-2023-large} has similarly found that LLMs may struggle with evaluating factual information in summarizations. Judge LLMs can even be swayed by short phrases like ``informative'', or ``solution:'' injected into evaluated texts \cite{raina-etal-2024-llm, zhao2025tokenfoolllmasajudge}.

Patterns of biases towards other aspects of the texts, e.g. authorship and length, are also found in both judging outputs of different language models and between human and model outputs. \citet{laurito2024aiaibiaslarge} found that LLMs often favor outputs of other LLMs over human outputs, and between different AI output texts. \citet{hu2024explaininglengthbiasllmbased} found that longer generated responses are preferred by LLM judges than shorter responses. \citet{panickssery2024favorown} and \citet{wataoka2024favorown2} found that LLMs also recognize and favor texts produced by the same model over other models. (i.e. GPT 4o-mini favor outputs of GPT 4o-mini over that of other LMs). Self-favoritism and position bias are also reported by \citet{zheng2023judgingllmasajudgemtbenchchatbot} and \citet{li-etal-2024-split}, with the former suggesting several mitigation strategies for position bias including a swapping operation and few-shot prompting and the latter proposing a split-and-merge approach to align semantically similar sections of evaluated texts. Large model evaluators' bias towards other models extends to other domains like images as well \cite{taesiri2025understandinggenerativeaicapabilities}.

\begin{table*}[ht]
\centering\resizebox{1.8\columnwidth}{!}{
    \begin{tabular}{rrrrr}
    \toprule
    Model name & \# of parameters & Context length & Summarizer & Evaluator \\
    \midrule
    google/gemma-3-1B-it & 1B & 32k & $\checkmark$ & $\checkmark$ \\
    google/gemma-3-4B-it & 4.3B & 128k & $\times$ & $\checkmark$ \\
    google/gemma-3-12B-it & 12.2B & 128k & $\times$ & $\checkmark$ \\
    \midrule
    meta-llama/Llama-3.2-3B-Instruct & 3.21B & 128k & $\times$ & $\checkmark$ \\
    meta-llama/Meta-Llama-3-8B-Instruct & 8.03B & 8k & $\checkmark$ & $\checkmark$ \\
    \midrule
    microsoft/Phi-4-mini-instruct & 3.84B & 128k & $\checkmark$ & $\checkmark$ \\
    mistralai/Mistral-7B-Instruct-v0.3 & 7.25B & 8k & $\checkmark$ & $\checkmark$ \\
    Qwen/Qwen3-8B & 8.19B & 32k & $\times$ & $\checkmark$ \\
    GPT-4o mini & 8B? & 128k & $\checkmark$ & $\checkmark$  \\
    \bottomrule
    \end{tabular}
}
\caption{Models used in the current work. Model names are Huggingface \texttt{repo\_id}s (excluding GPT-4o mini). ``Summarizer'' means the model was used to produce generated summaries; ``Evaluator'' means the model was used to evaluate summaries. GPT-4o mini's parameter count is unknown; however, \citet{gpt4omini-estimate-1} estimates 8 billion, citing \citet{gpt4omini-estimate-2}. We couldn't confirm this figure.}
\label{tab:model-info}
\end{table*}

\section{Methodology}\label{sec:methodology}
\subsection{Experimental Setup}\label{sec:dsets-llm-sums}

\begin{figure}[t!]
\begin{formal}
\small
\tt{%
Read the following section of a long document and write a concise summary that captures its main points and key details in around 100 words. Output the summary text only and nothing else.\\
\begin{center}
[original text]
\end{center}
}
\end{formal}
\caption{%
Prompt used for the LLMs to generate initial summaries to be evaluated
}
\label{fig:LLM-summary-prompt}
\end{figure}

\begin{figure}[t!]
\begin{formal}
\small
\tt{%
Given the original text below, along with indexed summaries, please evaluate the summaries and output the name of the best summary. Output the exact name only and nothing else.\\
original text:\\
\begin{center}
[original text]\\
\end{center}
summary\_1\\
\begin{center}
[first summary text]\\
\end{center}
summary\_2\\
\begin{center}
[second summary text]\\
\end{center}
}
\end{formal}
\caption{%
Prompt used for the LLMs to generate initial summaries to be evaluated.
}
\label{fig:LLM-eval-prompt}
\end{figure}

 We use test sets from \mbox{WikiSum} \cite{wikisum} and \mbox{CNN\_DailyMail} \cite{cnndailymail1, cnndailymail2}, covering a diverse range of topics for summarization. Both datasets contain 2,000 articles and their human written summaries in the test set. We use \mbox{Phi-4-mini-instruct} \cite{phi4mini}, \mbox{Mistral-7B-Instruct-v0.3} \cite{mistral7b}, \mbox{GPT-4o mini} \cite{chatgpt4}, and variants of Gemma \cite{gemma3} and LLaMA \cite{llama3}, covering parameter counts from 1 billion to 12 billion. The models are decoder-only transformers. A summary of tested models can be found in Table~\ref{tab:model-info}. During experiments, model temperatures are left at 0.7. For an LLM-generated summary, we use the average of BLEU-1, BLEU-4, ROUGE-1, and ROUGE-2 to score its similarity to the human summary. These four metrics cover recall- and precision-oriented scores and a range of n-gram lengths to capture keyword and short phrases matches.

We first obtain LLM summaries (``generated'' summaries) using the prompt in Figure~\ref{fig:LLM-summary-prompt}, where ``original text'' is the article in each of the datasets. We then prompt an evaluator model to judge pairs of summaries given the original texts using the prompt format in Figure~\ref{fig:LLM-eval-prompt}. Here, LLM-generated and human-generated ``ground truth'' summaries are assigned to either \texttt{[first summary text]} or \texttt{[second summary text]} as described later in Section~\ref{sec:control-for-length}. We keep no conversation history with LLMs so that they are not influenced by past summarizations or evaluations.

While our instruction to the judge LLMs is to only respond with the name of the summary, models occasionally return answers that quote their choices. We perform string matching to recover some judgments from these texts.

\subsection{Controlling for Length and Order Bias}\label{sec:control-for-length}
ROUGE and BLEU scores are affected by the length difference between reference and input texts, as longer input texts can simply include duplicates of segments in the reference text to inflate the scores. At the same time, LLM judges can be biased towards longer texts as discussed in Section~\ref{sec:related}. To reduce the effect of length bias, we control for summary length by filtering each dataset so that the reference (human-generated) summaries are between 95 and 105 space-delimited words long. Words are counted with space delimiters instead of using a particular tokenization algorithm because each model may tokenize texts differently. In the prompt in Figure~\ref{fig:LLM-summary-prompt}, we instruct LLMs to output 100 words to match lengths of ground truth summaries. After filtering, the CNN\_DailyMail dataset contains 286 articles and the WikiSum dataset contains 276 articles.

\begin{figure}[htp!]
    \centering
    \includegraphics[width=1\columnwidth]{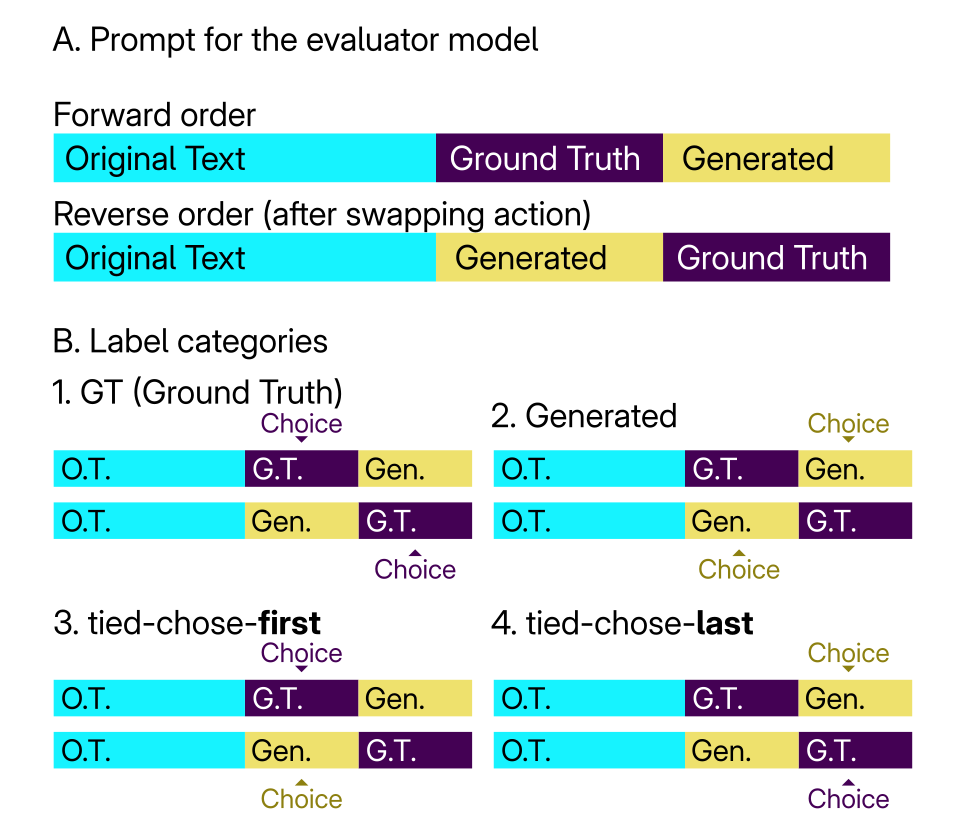}
    \caption{Visual representation of evaluator choice labels. ``GT'' means the evaluator chooses the ground truth summary in both orders; ``Generated'' means the evaluator chooses the LLM-generated summary in both orders. ``Tied-chose-first'' means the evaluator chooses the first presented summary in both orders, and ``Tied-chose-last'' means the evaluator chooses the last presented summary in both orders.}
    \label{fig:labels-cats}
\end{figure}

To reduce the effects of ordering bias and self-preferential bias as mentioned in Section~\ref{sec:related}, we perform evaluations with summaries presented in both orders. The evaluator LLM choices are accepted if they are consistent when summaries are presented in both orders and mark choices as ``tied'' if they are different depending on summary order.
Tied choices are further broken down by their ordering preference, i.e. if a model chooses the first summary for both orders, the tie is marked as ``tied-chose-first'', and if it chooses the last summary for both orders, the tie is marked as ``tied-chose-last''. We provide a visual representation of these categories in Figure~\ref{fig:labels-cats}.

\subsection{Extending the Range of Similarity Scores}\label{sec:rephrase}

After obtaining summaries and scoring their similarity with human summaries, we observed that the generated summaries had limited range for the averaged score, namely below 0.55. To get a fuller picture with wider range of similarity scores, we obtain additional LLM summaries by submitting ground truth summaries and prompting models to rephrase and reorganize them, keeping longer expressions and segments intact (see Figure~\ref{fig:LLM-rephrase-prompt} for prompt). Since summaries with more long phrase overlaps would score higher for ROUGE and BLEU, the additional summaries extend the range of scores for similarity metrics. These summaries are treated as input summaries and passed to evaluator LLMs as described in Section~\ref{sec:dsets-llm-sums}, i.e. the evaluator prompt does not reveal that these summaries are rephrased from ground-truth summaries.

\begin{figure*}[htp]
    \centering
    \includegraphics[width=1\textwidth]{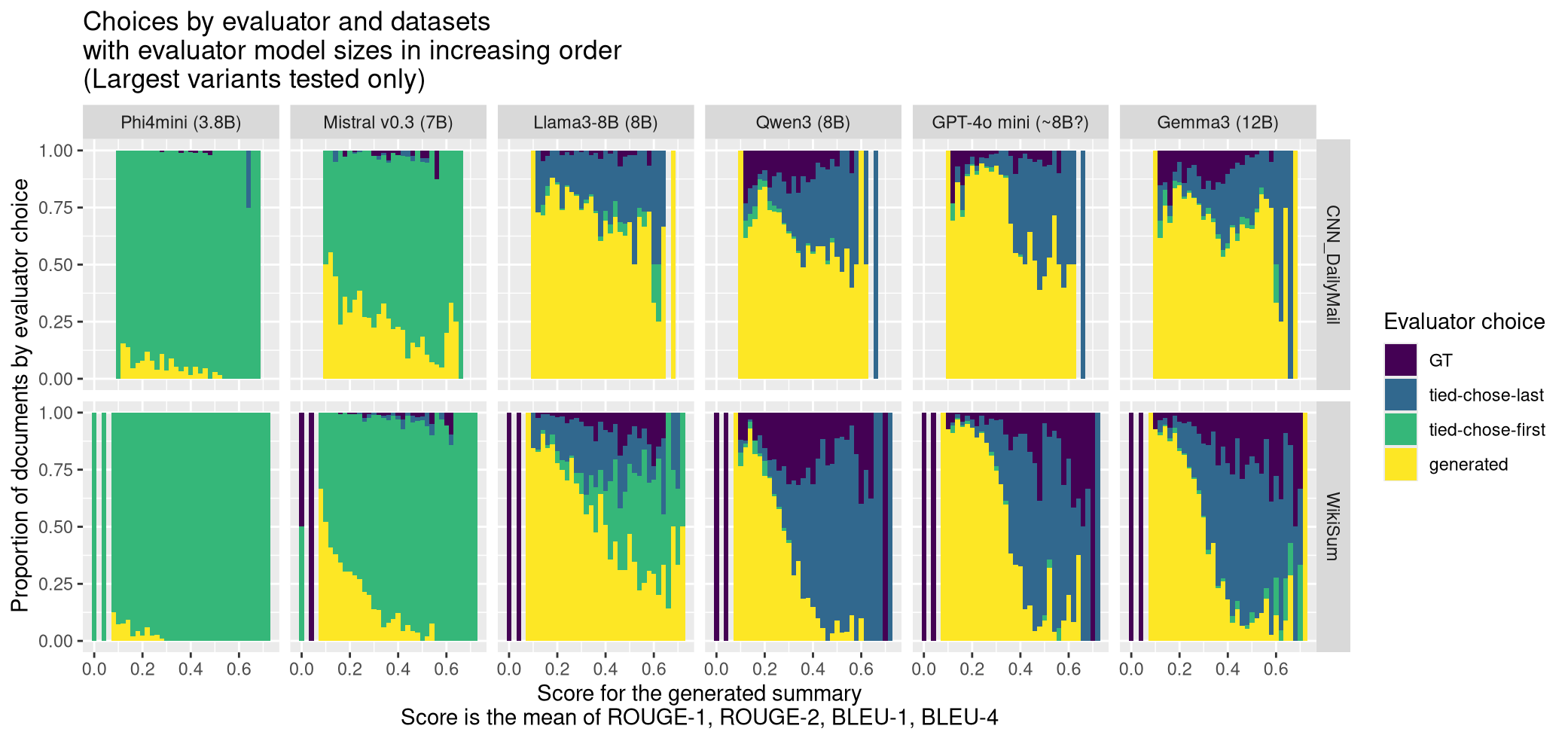}
    \caption{Proportion of documents where the evaluator chooses ground truth (\textbf{GT}), \textbf{generated} summaries, and when the evaluator \textbf{chose first}, \textbf{chose last} regardless of order, plotted against the score of the non-ground truth summaries. See Figure~\ref{fig:labels-cats} for a visual representation of evaluator choices. The score is the mean of ROUGE-1, ROUGE-2, BLEU-1, and BLEU-4. Here the generators (summarizers) are Gemma 3, Phi 4 mini, Mistral, Llama 3, and GPT-4o mini (i.e. no Qwen 3). For Llama 3 8B and Mistral, additional summaries are generated with different prompts to ascertain possible patterns in higher-scored summaries. 
    Further breakdowns for the variants Llama and Gemma can be found in Figure~\ref{fig:model-variants}.
    }
    \label{fig:stacked-gt-vs-gen}
\end{figure*}

\begin{figure*}[!h]
    \centering
    \includegraphics[width=0.9\textwidth]{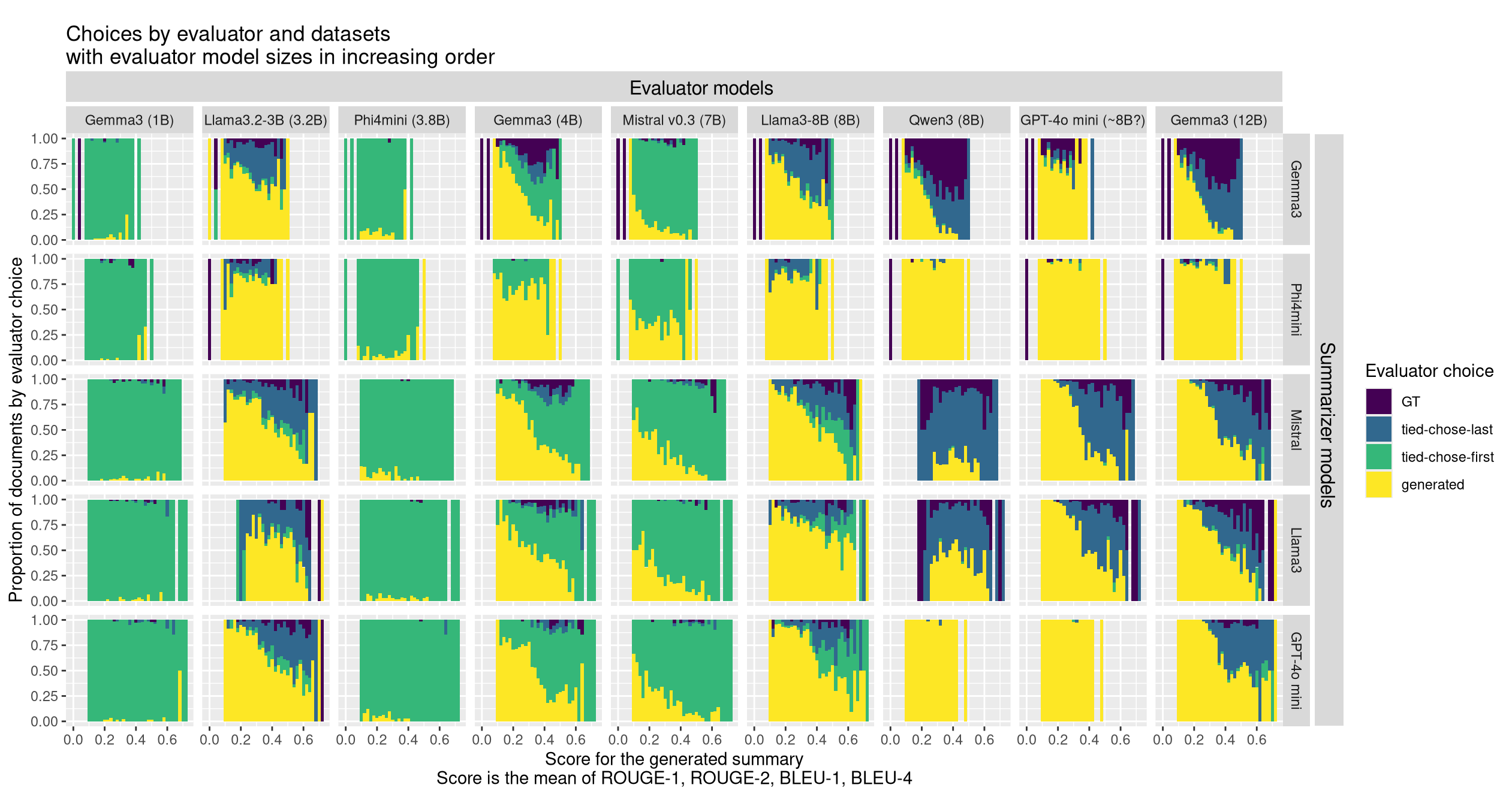}
    \caption{Alternative version of Figure~\ref{fig:stacked-gt-vs-gen}, where each row of the grid are the results with the same summarizer instead.}
    \label{fig:stacked-sum-vs-eval}
\end{figure*}

\begin{figure}[H]
\begin{formal}
\small
\tt{%
Rephrase and reorganize this text in your own style, but retain as many long phrases in it as possible. Keep to the same length. Output your rephrased text and nothing else.\\
\begin{center}
[ground truth summary]\\
\end{center}
}
\end{formal}
\caption{%
Prompt used for the LLMs to generate additional, higher-scoring summaries.
}
\label{fig:LLM-rephrase-prompt}
\end{figure}

\begin{figure*}[!h]
    \centering
    \includegraphics[width=0.7\textwidth]{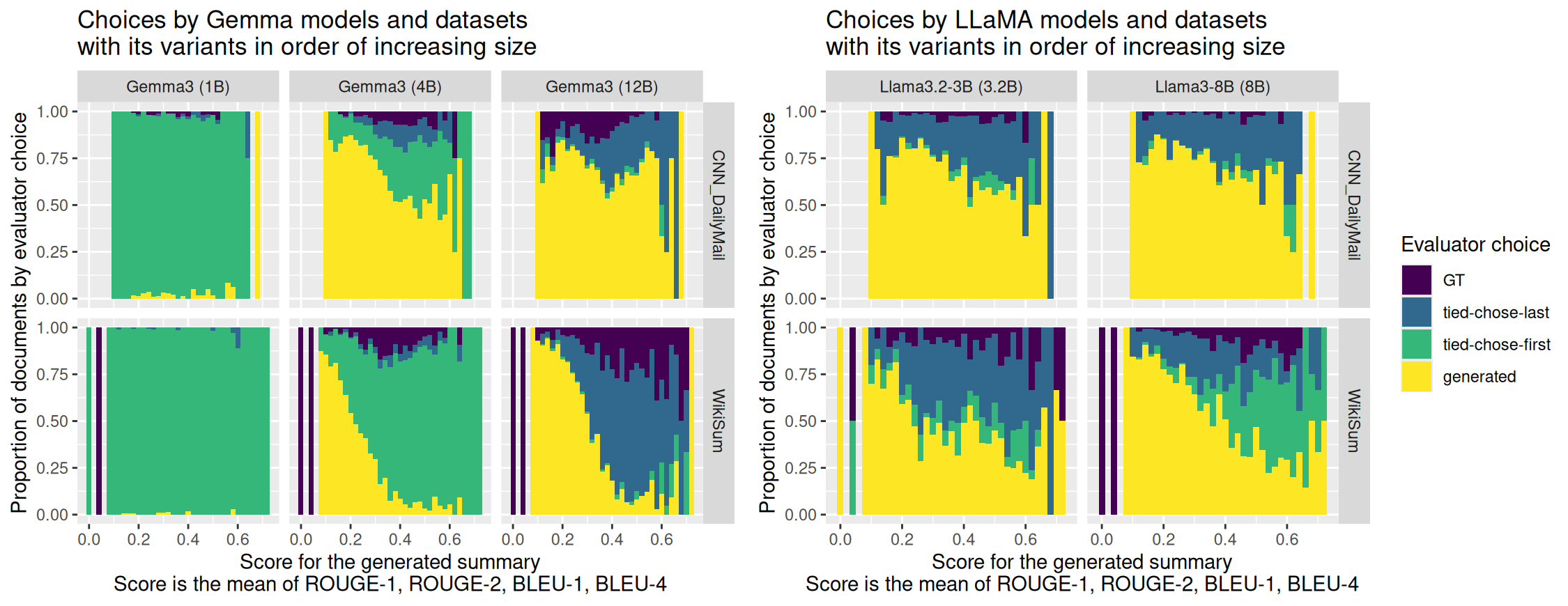}
    \caption{Alternative version of Figure~\ref{fig:stacked-gt-vs-gen}, showing breakdowns of variants of Llama and Gemma.}
    \label{fig:model-variants}
\end{figure*}

\section{Results \& Discussion}

Since model choices are discrete and only consist of 4 categories (excluding ``other''), we present the results in histograms to better visualize model preferences. For each graph in Figures~\ref{fig:stacked-gt-vs-gen} and~\ref{fig:stacked-sum-vs-eval}, we show the proportion of model choices for each bin of values on the x-axis.

\textbf{RQ1: How do LLM judgments correlate with n-gram-based similarity metrics, and is that consistent across models of different sizes?}\label{sec:llm-to-human-correlations}

We first note that for all models, the human summary is rarely chosen as the better summary regardless of similarity to human summaries. For all models excluding variants and the smallest Gemma-3-1B-it we observe a pattern that AI-AI bias (i.e. LLM tending to choose generated summaries over human summaries) is more prominent when generated summaries are less like the human-written summaries, i.e. fewer n-gram overlaps; see Figure~\ref{fig:stacked-gt-vs-gen} where the proportion of ``generated'' choices is larger towards the left of the subfigures. We see this tendency for LLM judges even when judging summaries produced by smaller models. For example, the preference patterns for the 12B-parameter Gemma 3 judge and the 8B Mistral judge are similar for summaries produced by the 1B Gemma 3 (see Figure~\ref{fig:stacked-sum-vs-eval}), even though the position biases of the judges are different, namely that Gemma 3 (12B) tend to choose last-presented summary while Mistral (7B) prefer the first-presented summary. It is worth noting that for most of the models tested the bias towards generated summaries diminish well before the average scores approaches 1. For example, Mistral's (8B) preference frequency for LLM summaries drops below 25\% for mean scores above 0.5. In other words, a significant non-overlap is required for the bias towards generated summaries to show.

\textbf{RQ2: How does position bias interact with the judgment vs. similarity pattern?}

We observe that position bias is more prominent when the generated summaries are more like human-written summaries. In Figure~\ref{fig:stacked-gt-vs-gen} this is represented by larger proportion of ``tied'' summaries towards the right of each subfigure. Models with more parameters tend to prefer the last-presented summaries while models with fewer parameters prefer the first-presented summaries. We note that for variously sized models we have tested, even though models may have different patterns of position bias, the pattern for the preference of ``generated'' summaries persists.

We find that LLM preferences towards generated summaries are similar for models with the same architecture above a certain parameter count, as exemplified in the two very different sizes of LLaMA models and the two biggest Gemma models tested.

\section{Conclusion}
In this work we have investigated the relationship between the human-machine text similarities and judge LLM's preference, specifically for the summarization task. We find that LLMs prefer generated summaries, and this preference is more prominent when generated and human summaries are less similar as measured by overlap metrics. This preference extends to summaries generated by smaller models, e.g. 1 billion parameters. We also find that this preference exists regardless of the type of position bias (preferring the first or last summary in both orders) present in that model. LLM bias towards not just their own but other LLMs over human texts points to a possible stylistic marker in LLM output text that is present even with varied training techniques and training data, which could be useful in contexts like LLM detection but unproductive in scenarios like LLM automatic evaluations. At the same time, judging bias displayed by this diverse set of models also could reveal areas for improvement in future LLM training and LLM-as-a-judge frameworks.

\section{Limitations}
The main limitation of this study is the scope for the independent variables for LLM bias. We have exclusively focused on judge LLM bias vs. degree of short phrases overlap metrics as a crude approximation for similarity with human-generated texts. Future studies could benefit from investigating many more potential metrics as the independent variable. Additionally, while we have used 9 models to generate summaries for evaluation, only one reference text was used to test judge LLM bias and to calculate overlap metrics. Future research may obtain multiple diverse \textit{human}-written summaries for more robust results for LLM bias patterns. In controlling for summary length, we have limited the findings to the subset of the datasets where human summaries are between 95 and 105 space-delimited words, and further work can expand on this range for both human and machine generated summaries. Finally, We have not considered adversarial examples, which may disrupt bias patterns.

\newpage
\section{References}\label{sec:reference}

\bibliographystyle{lrec2026-natbib}
\bibliography{lrec2026-example}

\label{lr:ref}
\bibliographystylelanguageresource{lrec2026-natbib}
\bibliographylanguageresource{languageresource}

\end{document}